\documentclass[sigconf]{acmart}
\usepackage{multirow}
\usepackage{url}
\usepackage{array}
\usepackage{subfigure}
\usepackage{balance}
\usepackage[misc]{ifsym}
\usepackage{graphicx}
\usepackage{enumitem}
\usepackage{makecell}
\AtBeginDocument{%
  \providecommand\BibTeX{{%
    \normalfont B\kern-0.5em{\scshape i\kern-0.25em b}\kern-0.8em\TeX}}}

\newcommand{\sports}{\textsc{SportsSum }}
\newcommand{\dataset}{\textsc{SportsSum2.0 }}
\copyrightyear{2021} 
\acmYear{2021} 
\setcopyright{acmcopyright}\acmConference[CIKM '21]{Proceedings of the 30th ACM International Conference on Information and Knowledge Management}{November 1--5, 2021}{Virtual Event, QLD, Australia}
\acmBooktitle{Proceedings of the 30th ACM International Conference on Information and Knowledge Management (CIKM '21), November 1--5, 2021, Virtual Event, QLD, Australia}
\acmPrice{15.00}
\acmDOI{10.1145/3459637.3482188}
\acmISBN{978-1-4503-8446-9/21/11}

\begin{document}

\fancyhead{}
\title{\textsc{SportsSum2.0}: Generating High-Quality Sports News from Live Text Commentary}

\author{
Jiaan Wang$^{\dagger*}$, Zhixu Li$^{\ddagger*(\textrm{\Letter})}$, Qiang Yang$^{\clubsuit}$, Jianfeng Qu$^{\dagger}$, Zhigang Chen$^{\spadesuit\heartsuit}$ \\ Qingsheng Liu$^{\diamondsuit}$, and Guoping Hu$^{\spadesuit}$
}
\makeatletter
\def\authornotetext#1{
 \g@addto@macro\@authornotes{%
 \stepcounter{footnote}\footnotetext{#1}}%
}
\makeatother

\authornotetext{The first two authors made equal contributions to this work.}

\affiliation{%
  \institution{$^{\dagger}$ School of Computer Science and Technology, Soochow University, Suzhou, China}
  \country{}
}
\affiliation{%
  \institution{$^{\ddagger}$ Shanghai Key Laboratory of Data Science, School of Computer Science, Fudan University}
  \country{}
}
\affiliation{%
  \institution{$^{\clubsuit}$ King Abdullah University of Science and Technology \quad $^{\heartsuit}$ iFLYTEK Research, Suzhou}
  \country{}
}

\affiliation{%
  \institution{$^{\spadesuit}$ State Key Laboratory of Cognitive Intelligence, iFLYTEK Research \quad $^{\diamondsuit}$ Anhui Toycloud Technology}
  \country{}
}

\email{jawang1@stu.suda.edu.cn, zhixuli@fudan.edu.cn}

\begin{abstract}
Sports game summarization aims to generate news articles from live text commentaries.
A recent state-of-the-art work, \textsc{SportsSum}, not only constructs a large benchmark dataset, but also proposes a two-step framework. Despite its great contributions, the work has three main drawbacks: 1) the noise existed in \textsc{SportsSum} dataset degrades the summarization performance; 2) the neglect of lexical overlap between news and commentaries results in low-quality pseudo-labeling algorithm; 3) the usage of directly concatenating rewritten sentences to form news limits its practicability.
In this paper, we publish a new benchmark dataset \textsc{SportsSum2.0}, together with a modified summarization framework.
In particular, to obtain a clean dataset, we employ crowd workers to manually clean the original dataset.
Moreover, the degree of lexical overlap is incorporated into the generation of pseudo labels.
Further, we introduce a reranker-enhanced summarizer to take into account the fluency and expressiveness of the summarized news.
Extensive experiments show that our model outperforms the state-of-the-art baseline.
\end{abstract}




\maketitle

{
  \medskip\small\noindent{\bfseries ACM Reference Format:}\par\nobreak
  \noindent\bgroup\def\\{\unskip{}, \ignorespaces}{Jiaan Wang, Qiang Yang, Jianfeng Qu, Zhixu Li, Zhigang Chen, Qingsheng Liu, Guoping Hu}\egroup. 2021. \textsc{SportsSum2.0}: Generating High-Quality Sports News from Live Text Commentary. In \textit{Proceedings of the 30th ACM Int'l Conf. on Information and Knowledge Management (CIKM '21), November 1--5, 2021, Virtual Event, Australia}\textit{.} ACM, New York, NY, USA, \ref{TotPages}~pages. https://doi.org/10.1145/3459637.3482188
  }

\section{Introduction}
\begin{figure}[t]
\centerline{\includegraphics[width=0.40\textwidth]{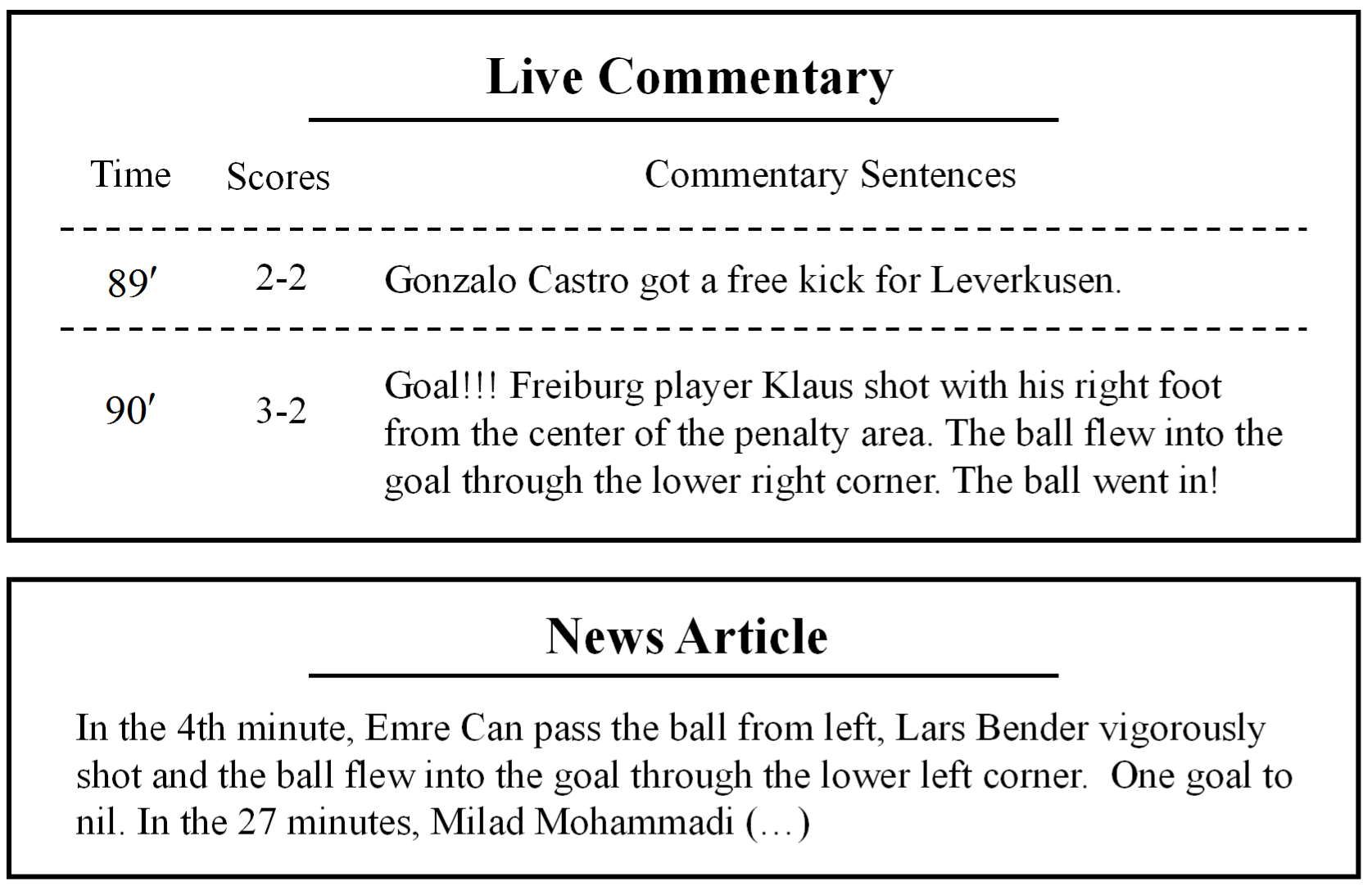}}
\caption{An example of Sports Game Summarization.}
\label{example}
\end{figure}

Text Summarization aims at compressing the original document into a shorter text while preserving the main ideas~\cite{rush-etal-2015-neural,chopra-etal-2016-abstractive,Nallapati2016AbstractiveTS,See2017GetTT,chen-bansal-2018-fast}.
%
A special text summarization task in sports domain is Sports Game Summarization, as the example shown in Fig.~\ref{example}, which focuses on generating news articles from live commentaries~\cite{zhang-etal-2016-towards}. Obviously, this task is more challenging than conventional text summarization for two reasons: First, the length of live commentary document often reaches thousands of tokens, which is far beyond the capacity of mainstream PLMs (e.g., BERT, RoBERTa, etc.); Second, there are different text styles between commentaries and news. Specifically, commentaries are more colloquial than news.

Sports game summarization has gradually attracted attention from researchers due to its practical significance.
Zhang et al.~\cite{zhang-etal-2016-towards} pioneer this task and construct the first dataset with only 150 samples. Another dataset created for the shared task at NLPCC 2016 contains 900 samples~\cite{Wan2016OverviewOT}. Although these datasets promote the research to some extent, they cannot further fit more sophisticated models due to their limited scale.
Additionally, early literature~\cite{zhang-etal-2016-towards,Zhu2016ResearchOS,Yao2017ContentSF,Liu2016SportsNG} on these datasets mainly explores different strategies to select important commentary sentences to form news directly, but ignores the different text styles between commentaries and news.
In recent, Huang et al.~\cite{Huang2020GeneratingSN} present \textsc{SportsSum}, the first large-scale sports game summarization dataset with 5,428 samples. This work also proposes a two-step model which first \textbf{selects} important commentary sentences, and then \textbf{rewrites} each selected sentence to a news sentence so as to form news.
In order to provide training data for both \textbf{selector} and \textbf{rewriter} of the two-step model, a pseudo-labeling algorithm is introduced to find for each news sentence a corresponding commentary sentence according to their timeline information as well as semantic similarities.

Given all the existing efforts, this task is still not fully exploited in the following aspects:
(1) The existing datasets are limited in either scale or quality. According to our observations on \textsc{SportsSum}, more than 15\% of samples have noisy sentences in the news articles due to its simple rule-based data cleaning process;
(2) The pseudo-labeling algorithm used in existing two-step models only considers the semantic similarities between news sentences and commentary sentences but neglects the lexical overlap between them, which actually is an useful clue for generating the pseudo label;
(3) The existing approaches rely on direct stitching to constitute news with the (rewritten) selected sentences, resulting in low fluency and high redundancy problems due to each (rewritten) selected sentence is transparent to other sentences. 

Therefore, in this paper, we first denoise the \textsc{SportSum} dataset to obtain a higher-quality \textsc{SportsSum2.0} dataset.
Secondly, lexical overlaps between news sentences and commentary sentences are taken into account by our advanced pseudo-labeling algorithm.
Lastly, we extend the two-step framework~\cite{Huang2020GeneratingSN} to a novel reranker-enhanced summarizer, where the last step reranks the rewritten sentences w.r.t importance, fluency and redundancy.
We evaluate our model on the \dataset and \sports~\cite{Huang2020GeneratingSN} datasets. The experimental results show that our model achieves state-of-the-art performance in terms of ROUGE Scores. 

\begin{figure}[t]
\centerline{\includegraphics[width=0.40\textwidth]{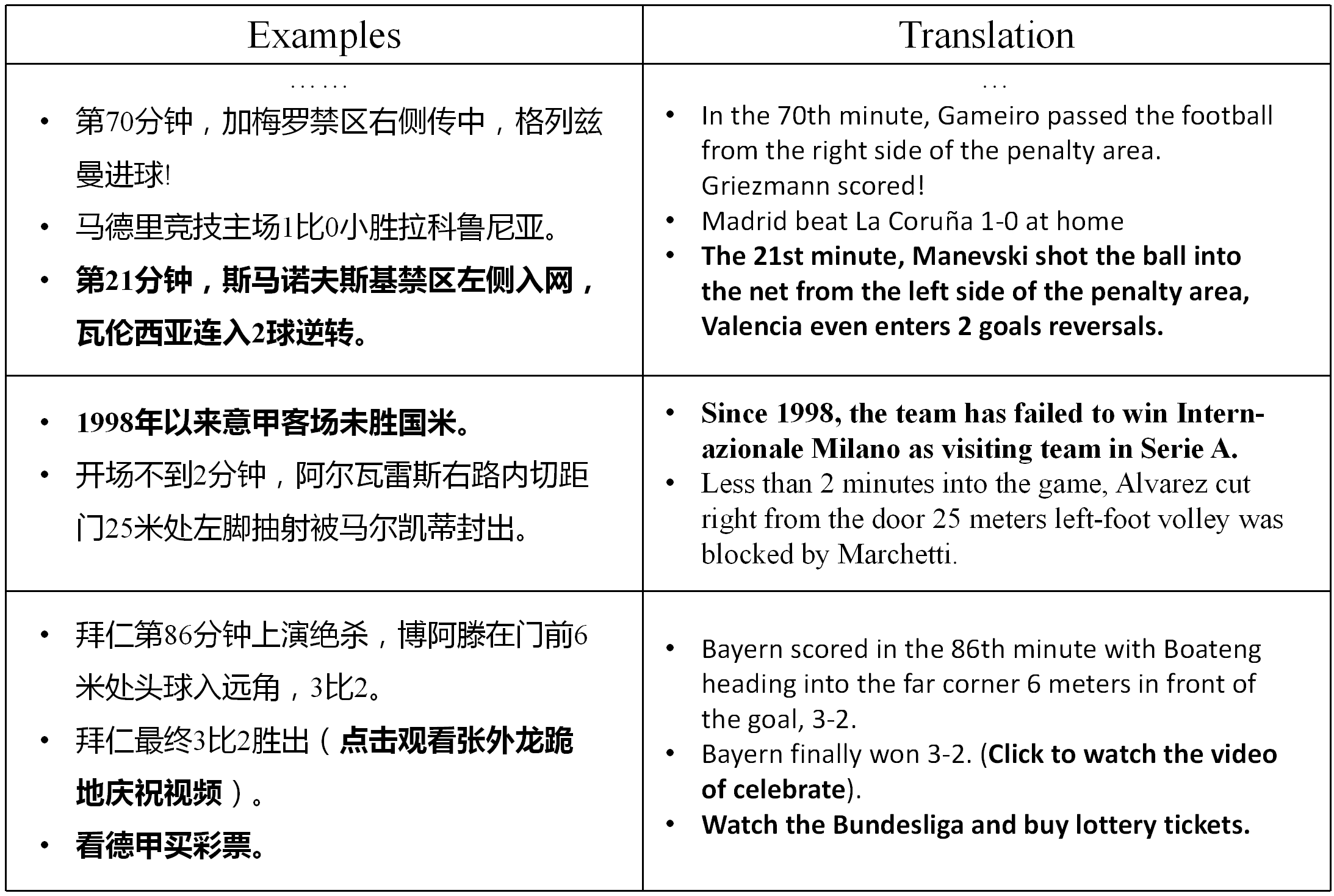}}
\caption{Noise existed in \sports Dataset. The first example contains the descriptions of other games while the second one has the history-related descriptions. The last case includes an advertisement and irrelevant hyperlink text.}
\label{noisy_samples}
\end{figure} 

\section{Data Construction}
In this section, we first analyze the existing noises in \textsc{SportsSum}, and then introduce the details of manual cleaning process. Lastly, we show the statistics of \textsc{SportsSum2.0}.

\vspace{1ex}
\noindent\textbf{Noise Analysis.}
\textsc{SportsSum}~\cite{Huang2020GeneratingSN} is the only public large-scale sports game summarization dataset with 5,428 samples crawled from Chinese sports website. During the observation of \textsc{SportsSum}, we find more than 15\% of \emph{news articles} have noisy sentences. Specifically, we divide these noises into three classes and show an example for each class in Fig.~\ref{noisy_samples}:
\begin{itemize}[leftmargin=*,topsep=0pt]
\item Descriptions of other games: a news website may contain news articles of multiple games, which is neglected by \textsc{SportsSum} and resulted in 2.2\% (119/5428) of news articles include descriptions of other games. 
\item Descriptions of history: There are many sports news describe the matching history at the beginning, which cannot be inferred from the commentaries. To alleviate this issue, \sports adopts a rule-based method to identify the starting keywords (e.g., ``at the beginning of the game'') and remove the descriptions before the keywords. However, about 4.6\% (252/5428) of news articles cannot be correctly disposed by this method.
\item Advertisements and irrelevant hyperlink text: We find that about 9.8\% (531/5428) of news articles in \sports have such noise.
\end{itemize}

\vspace{1ex}
\noindent\textbf{Manual Cleaning Process.}
In order to reduce the noise of \emph{news articles}, we design a manual cleaning process. For a given news article, we first remove the descriptions related to other games, and then delete advertisements and irrelevant hyperlink text. Finally, we discard the descriptions of history.

\vspace{1ex}
\noindent\textbf{Annotation Process.}
We recruit 7 master students to perform the manual cleaning process. In order to help the annotators fully understand our manual cleaning process, every annotator is trained in a pre-annotation process. All the cleaning results are checked by another two data experts. The manual cleaning process costs about 200 human hours in total. After manual cleaning, we discard 26 bad cases which do not contain the descriptions of current games and finally obtain 5402 human-cleaned samples. 

\vspace{1ex}
\noindent\textbf{Statistics.}
Table ~\ref{table:statistic} shows statistics of \textsc{SportsSum2.0} and \textsc{SportsSum}. The average length of commentary documents in \textsc{SportsSum2.0} is slightly different from the counterpart in \textsc{SportsSum} due to the 26 removed bad cases.

\begin{table}[t]
\setlength{\belowcaptionskip}{5pt}
\centering
\resizebox{0.40\textwidth}{!}{
  \begin{tabular}{|c|cc|cc|}
  \hline
  \multirow{2}{*}{Source} & \multicolumn{2}{c|}{\textsc{SportsSum2.0}} & \multicolumn{2}{c|}{\textsc{SportsSum}} \\
                                  & Commentary       & News        & Commentary        & News          \\ \hline
  Avg. \#chars           & 3464.31          & 771.93      & 3459.97           & 801.11        \\
  Avg. \#words            & 1828.56          & 406.81      & 1825.63           & 427.98        \\
  Avg. \#sent.            & 194.10           & 22.05       & 193.77            & 23.80         \\ \hline
  \end{tabular}
}
\caption{The statistics of \textsc{SportsSum2.0} and \textsc{SportsSum}.}
\label{table:statistic}
\vspace{-7pt}
\end{table}

\section{Methodology}
As shown in Fig.~\ref{example}, the goal of sports game summarization is to generate sports news $R=\{r_{1},r_{2},...,r_{n}\} $ from a given live commentary document $C=\{(t_{1},s_{1},c_{1}),...,(t_{m},s_{m},c_{m})\}$ $(m \geq n)$. $r_{i}$ represents $i$-th news sentence and $(t_{j},s_{j},c_{j})$ is $j$-th commentary, where $t_{j}$ is the timeline information, $s_{j}$ denotes the current scores and $c_{j}$ is the commentary sentence.

\begin{figure}[t]
  \centerline{\includegraphics[width=0.45\textwidth]{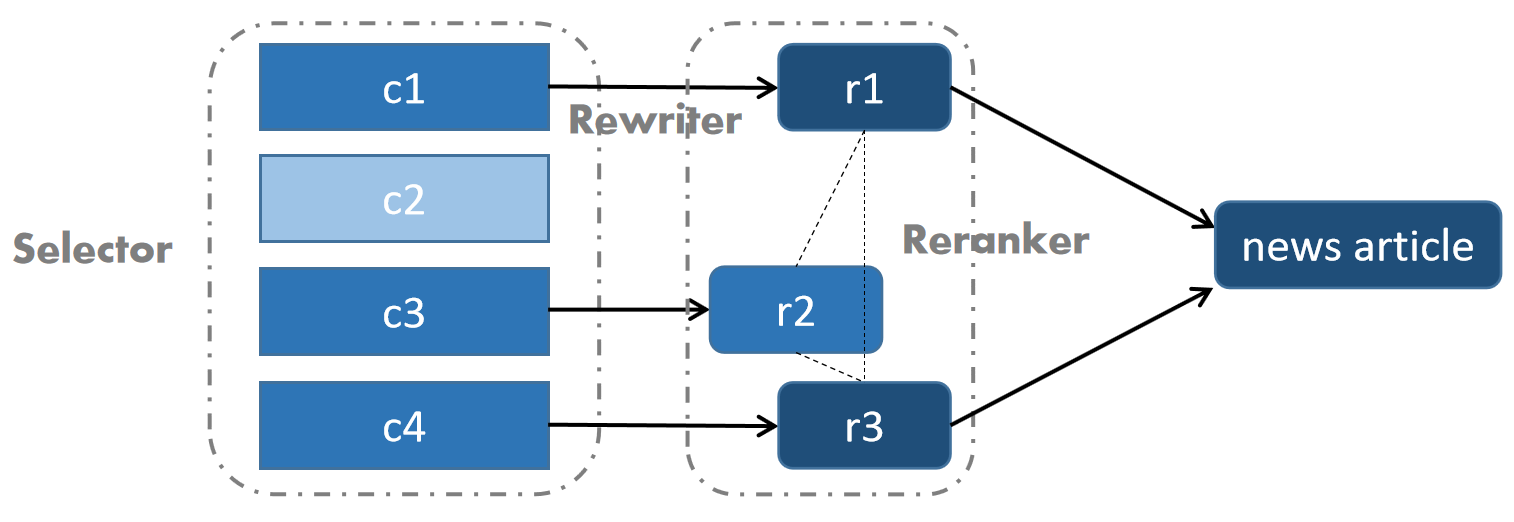}}
  \caption{The overview of the reranker-enhanced summarizer.}
  \label{fig:overview}
  \end{figure}

Fig.~\ref{fig:overview} shows the overview of our reranker-enhanced summarizer which first learns a \textbf{selector} to extract important commentary sentences, then uses a \textbf{rewriter} to convert each selected sentence to a news sentence. Finally,  a \textbf{reranker} is introduced to generate a news article based on the rewritten sentences.

\subsection{Pseudo-Labeling Algorithm}
To train the selector and rewriter, we need labels to indicate the importance of commentary sentences and their corresponding news sentences. Following \textsc{SportsSum}\cite{Huang2020GeneratingSN}, to obtain the labels, we \emph{map each news sentence to its commentary sentence} through a pseudo-labeling algorithm that considers both timeline information and similarity metrics.
Though there is no explicit timeline information of news sentences, but most of sentences start with ``in the n-th minutes'' which indicates its timeline information.

For each news sentence $r_{i}$, we first extract time information $h_{i}$ if possible, and then construct a candidate commentaries set $C^{(i)}$, where the timeline information of each commentary belongs to $[h_{i},h_{i}+3]$. Lastly, we select one commentary sentence with the highest similarity with $r_{i}$ from $C^{(i)}$, and form a pair of a mapped commentary sentence and a news sentence.

The similarity function used by Huang et al.~\cite{Huang2020GeneratingSN} is BERTScore~\cite{Zhang2020BERTScoreET}, which only considers the semantic similarity between news and commentaries but neglects the lexical overlap between them. However, the lexical overlap actually is a useful clue, e.g., the matching probability of a news sentence and a commentary sentence will be greatly improved if the same entity mentions appear both in these two sentences. Therefore, we decide to consider both semantics and lexical overlap in similarity function:
\begin{equation}
\label{similarity_function}
S(r_{i},c_{j}) = \lambda B(r_{i},c_{j}) + (1-\lambda) R(r_{i},c_{j})
\end{equation}

The similarity function, i.e., $S(\cdot, \cdot)$ is calculated by linearly combining BERTScore function, i.e., $B(\cdot, \cdot)$ and ROUGE score function, i.e., $R(\cdot, \cdot)$. The coefficient $\lambda$ is a hyper parameter.

With the above process, we finally obtain a large number of pairs of a mapped commentary sentence and a news sentence, which can be used for training selector and rewriter.

\begin{table*}[t]
\setlength{\belowcaptionskip}{5pt}
  \centering
  \resizebox{0.80\textwidth}{!}
  {
    \centering
    \begin{tabular}{cclcccccc}
      \cline{1-9}
      \multirow{2}{*}{Method}   & \multirow{2}{*}{\#}                                                     & \multirow{2}{*}{Model} & \multicolumn{3}{c}{\dataset}                                   & \multicolumn{3}{c}{\sports}                                      \\
                                                                                     &  &                      & ROUGE-1              & ROUGE-2              & ROUGE-L              & ROUGE-1              & ROUGE-2              & ROUGE-L              \\ \cline{1-9}
      \multirow{2}{*}{Extractive Models}    & 1                                         & TextRank               & 20.53                & 6.14                 & 19.64                & 18.37                & 5.69                 & 17.23                \\
                                                         & 2                            & PacSum                 & 23.13                & 7.18                 & 22.04                & 21.84                & 6.56                 & 20.19                \\ \cline{1-9}
      \multirow{2}{*}{Abstractive Models}                & 3                            & Abs-LSTM                   & 31.14                & 11.22                & 30.36                & 29.22                & 10.94                & 28.09                \\
                                                          & 4                           & Abs-PGNet                  & 35.98                & 13.07                & 35.09                & 33.21                & 11.76                & 32.37                \\ \cline{1-9}
      \multirow{5}{*}{\makecell[c]{Two Step Framework \\ (Selector + Rewriter)}}       & 5               & SportsSUM$^{\dagger}$              & 44.73                & 18.90                & 44.03                & 43.17                & 18.66                & 42.27                \\
                                                                               & 6      & PGNet$^{*}$                  & 45.13                & 19.13                & 44.12                & 44.12                & 18.89                & 43.23                \\
                                                                                & 7     & mBART$^{*}$                  & 47.23                & 19.27                & 46.54                & 46.43                & 19.54                & 46.21                \\
                                                                                & 8     & Bert2bert$\dagger$              & 46.85                & 19.24                & 46.12                & 46.54                & 19.32                & 45.93                \\
                                                                                & 9     & Bert2bert$^{*}$              & 47.54                & 19.87                & 46.99                & 47.08                & 19.63                & 46.87                \\ \cline{1-9}
      \multirow{4}{*}{\makecell[c]{Reranker-Enhanced Summarizer \\ (Selector + Rewriter + Reranker)}} & 10 & PGNet$^{*}$                  & 46.23                & 19.41                & 45.37                & 45.54                & 19.02                & 45.31                \\
                                                                            & 11         & mBART$^{*}$                  & 47.62                & 19.73                & 47.19                & 46.89                & \textbf{19.72}       & 46.53                \\
                                                                         & 12            & Bert2bert$^{\dagger}$             & 47.32       & 19.33       & 47.01       & 46.14      & 19.32                & 45.53       \\                                                                                    & 13         & Bert2bert$^{*}$             & \textbf{48.13}       & \textbf{20.09}       & \textbf{47.78}       & \textbf{47.61}       & 19.65                & \textbf{47.49}       \\ \cline{1-9}                                               
      \end{tabular}
  }
  \caption{Experimental results on \textsc{SportsSum2.0} and \textsc{SportsSum}. The models with $^{*}$ denotes they use our advanced pseudo-labeling algorithm while $^{\dagger}$ indicates the models utilize the original pseudo-labeling algorithm~\cite{Huang2020GeneratingSN}.}
  \label{table:result}
  \vspace{-2em}
\end{table*}

\subsection{Reranker-Enhanced Summarizer}
Our reranker-enhanced summarizer extends the two-step model~\cite{Huang2020GeneratingSN} to consider the importance, fluency and redundancy of the rewritten news sentences, as shown in Fig.~\ref{fig:overview}.
Particularly, we first select important commentary sentences, and then rewrite each selected sentence to a news sentence by a seq2seq model. However, the readability and fluency of these news sentences vary a lot. Some new sentences can be directly used in the final news while others are inappropriate due to the low fluency, which is one common problem in natural language generation, representing by repetition~\cite{See2017GetTT} and incoherence~\cite{bellec2017deep}.
So we use the reranker to filter out low fluent sentences while retaining high informative sentences and controlling the redundancy between the rewritten sentences.

\vspace{1ex}
\noindent\textbf{Selector.} Different from existing two-step model~\cite{Huang2020GeneratingSN}, which purely use TextCNN~\cite{Kim2014ConvolutionalNN} as selector and ignores the contexts of a commentary sentence, we design a context-aware selector which can capture the semantics well with a sliding window.
In detail, we train a binary classifier to choose important commentary sentences. When training, for each commentary $c_{i}$ in $C$, we assign a positive label if $c_{i}$ can be mapped with a news sentence by the pseudo-labeling algorithm. Otherwise, we give a negative label.
Here, a RoBERTa~\cite{Liu2019RoBERTaAR} is employed to extract the contextual representation of commentary sentences.
we first tokenize the commentary sentences. Then we concatenate the target commentary sentence and its partial context with special tokens as \texttt{[CLS] commentary1 [SEP] commentary2 [SEP] ... [SEP] commentaryN [SEP]}. the target commentary sentence in the middle of the whole sequence, and the sequence is limited to 512 tokens.
For prediction, the sentence embedding is obtained by averaging the token embedding of the target commentary sentence, and then is followed by a sigmoid classifier.
The cross-entropy loss is used as the training objective for selector.

\vspace{1ex}
\noindent\textbf{Rewriter.} Each selected commentary sentence first concatenates with its timeline information, and then rewrites to a news sentence through the seq2seq model. To be more specific, we choose the following three seq2seq models:

\begin{itemize}[leftmargin=*,topsep=0pt]
\item \emph{Pointer-Generator Network}~\cite{See2017GetTT} is a popular abstractive text summarization model with copy mechanism and coverage loss.
\item \emph{Bert2bert}\footnote{\url{https://huggingface.co/transformers/model_doc/encoderdecoder.html}} is a seq2seq model, in which both encoder and decoder are initialized by BERT~\cite{Devlin2019BERTPO}.
\item \emph{BART}~\cite{Lewis2020BARTDS} is a denoising autoencoder seq2seq model, which achieves state-of-the-art results on many language generation tasks. Since there is no Chinese version of BART for public use, we choose its multilingual version, i.e., mBART~\cite{Liu2020MultilingualDP}.
\end{itemize}

\vspace{1ex}
\noindent\textbf{Reranker.} Although the rewritten sentences can convey the semantic to some extent, there still exist low fluent sentences, which leads to poor readability.
In view of the ability of Maximal Marginal Relevance (MMR)~\cite{carbonell1998use} on extracting the high informative and low redundant sentences, we decide to adopt this approach as reranker. Unfortunately, vanilla MMR cannot take fluency into account. So, we propose a variant MMR algorithm by incorporating the fluency of each rewritten sentence. 
\begin{equation}
\begin{split}
MMR(D, R)={\underset {d_{i}\in D-R}{argmax}}[\lambda_{1}info(d_{i})+\lambda_{2}flu(d_{i})\\
-\lambda_{3}\max_{d_{j}\in R} [sim(d_{i},d_{j})]]\ \  (\lambda_{1} + \lambda_{2} + \lambda_{3} = 1)
\end{split}
\end{equation}
\begin{equation}
  \label{equ:flu}
    flu(d_{i}) = 1 - perplexity(d_{i})/\eta
\end{equation}
where $D$ represents the whole news sentence set, $R$ denotes the selected news sentence set.
The $info(d_{i})$ is employed to calculate the importance of the news sentence. Note that, each commentary sentence has corresponding importance predicted by selector. So we directly use the importance of corresponding commentary sentence as $info(d_{i})$.
The $flu$ function in our MMR algorithm is utilized to indicate the fluency of a news sentence. We consider computing perplexity of the news sentence by GPT-2~\cite{Radford2019LanguageMA}. Equal~\ref{equ:flu} shows the details of $flu$ function.
The $sim(d_{i},d_{j})$ is BERTScore function.
We greedily select news sentences with the highest MMR score until the total length exceeds a pre-defined budget\footnote{We set the average length of the news articles as the budget.}.

\section{Experiments}
\subsection{Implementation Details}
We split SportsSum2.0\footnote{The dataset is available at \url{https://github.com/krystalan/SportsSum2.0}} into three sets: training (4803 games), validation (300 games), and testing (299 games) sets$\footnote{We keep the original splitting except for removed 26 ``bad cases".}$. We trained all models on one 32GB Tesla V100 GPU. Our reranker-enhanced summarizer is implemented based on RoBERTa-large (24 layers with 1024 hidden size), mBART (12 layers with 1024 hidden size) and GPT-2 of huggingface Transformers~\cite{wolf-etal-2020-transformers} with default settings. The learning rates of selector, PGNet, Bert2bert rewriter and mBART rewriter are 3e-5, 0.15, 2e-5 and 2e-5, respectively. For all models, we set the batch size to 32 and use Adamw optimizer. The coefficient $\lambda$ in our pseudo-labeling algorithm is 0.70. We set the $\lambda_{1}$, $\lambda_{2}$ and $\lambda_{3}$ to 0.6, 0.2 and 0.2 in our variant MMR.

\begin{figure}[t]
\centering
\subfigure[Info]{
  \includegraphics[width=0.20\linewidth]{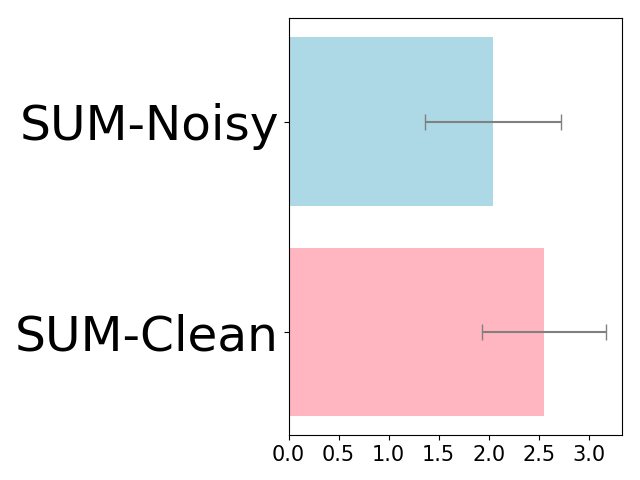}
}
\subfigure[Redundancy]{
  \includegraphics[width=0.20\linewidth]{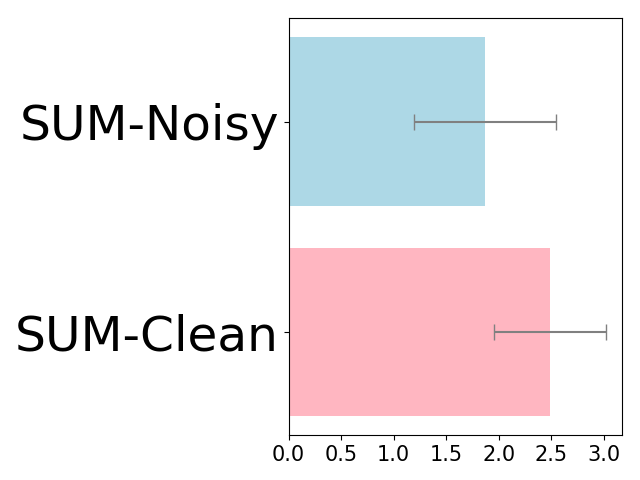}
}
\subfigure[Fluency]{
  \includegraphics[width=0.20\linewidth]{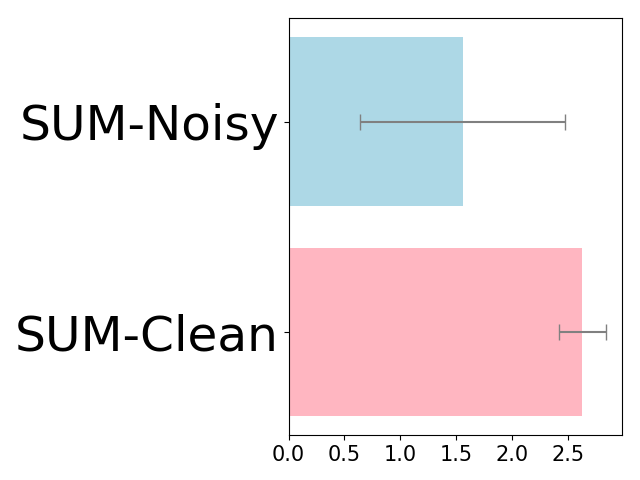}
}
\subfigure[Overall]{
  \includegraphics[width=0.20\linewidth]{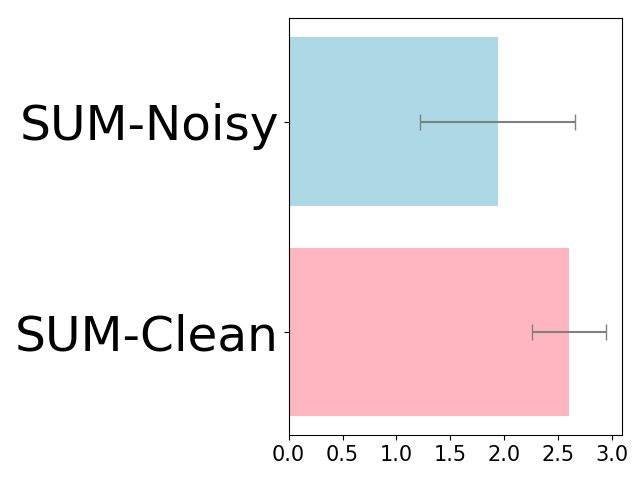}
}
\caption{Results on human evaluation.}
\label{fig:huamn_evaluation}
\end{figure}

\subsection{Experimental Results}
We compare our three-step models with several conventional models, including extractive summarization models, abstractive summarization models and two-step models in terms of ROUGE scores.

As shown in Table~\ref{table:result}, our reranker-enhanced models achieve significant improvement on both datasets.
TextRank~\cite{Mihalcea2004TextRankBO} and PacSum~\cite{zheng-lapata-2019-sentence} are extractive summarization models, which are limited by the different text styles between commentaries and news. Abs-PGNet~\cite{See2017GetTT} and Abs-LSTM (abstractive summarization models) achieve better performance since they can alleviate different styles issue, but still have the drawback of dealing with long texts.
SportsSUM~\cite{Huang2020GeneratingSN} is a two-step state-of-the-art model that performs better than the above models where the different text styles and long text issues are solved. 
Other two-step models enhance SportsSUM through improved pseudo-labeling algorithm, selector and rewriter. However, they still suffer from low fluency problem. Our best reranker-enhanced model outperforms SportsSUM by more than 2.8 and 3.5 points in the average of ROUGE scores on \textsc{SportsSum2.0} and \textsc{SportsSum}, respectively.
Additionally, the effectiveness of our advanced pseudo-labeling algorithm is proved by comparing row 8 to 9 or 12 to 13.

\subsection{Necessity of Manual Cleaning}
To further demonstrate the necessity of manual cleaning, we conduct a human evaluation on sports news generated by two reranker-enhanced summarizers trained on \textsc{SportsSum2.0} and \textsc{SportsSum}, respectively. We denote these two models as SUM-Clean and SUM-Noisy. Five postgraduate students are recruited and each one evaluates 100 samples for each summarizer. The evaluator scores generated sports news in terms of informativeness, redundancy, fluency and overall quality with a 3-point scale.

Fig.~\ref{fig:huamn_evaluation} shows the human evaluation results. SUM-Clean performs better than the SUM-Noisy on all four aspects, especially in fluency. This indicates that the summarizer trained on noisy data degrades the performance of game summarization, and it is necessary to remove the noise existed in the dataset.



\section{Conclusion}
In this paper, we study the sports game summarization on the basis of SportsSum. A high-quality dataset SportsSum2.0 is constructed by removing or rectifying noises. We also propose a novel pseudo-labeling algorithm based on both semantics and lexicons. Furthermore, an improved framework is designed to improve the fluency of rewritten news. Experimental results show the effectiveness of our model in the SportsSum2.0.

\begin{acks}
This research is supported by National Key R\&D Program of China (No. 2018-AAA0101900), the Priority Academic Program Development of Jiangsu Higher Education Institutions, National Natural Science Foundation of China (Grant No. 62072323, 61632016), Natural Science Foundation of Jiangsu Province (No. BK20191420), Suda-Toycloud Data Intelligence Joint Laboratory, and the Collaborative Innovation Center of Novel Software Technology and Industrialization.
\end{acks}

\bibliographystyle{ACM-Reference-Format}
\balance
\bibliography{sportssum}


\end{document}